# A versatile circuit for emulating active biological dendrites applied to sound localisation and neuron imitation

**Daniel J. Mannion**

The Alan Turing Institute, London, United Kingdom.
Correspondence E-mail: daniel.m@turing.ac.uk

## Abstract

Sophisticated machine learning struggles to transition onto battery-operated devices due to the high-power consumption of neural networks. Researchers have turned to neuromorphic engineering, inspired by biological neural networks, for more efficient solutions. While previous research focused on artificial neurons and synapses, an essential component has been overlooked: dendrites. Dendrites transmit inputs from synapses to the neuron's soma, applying both passive and active transformations. However, neuromorphic circuits replace these sophisticated computational channels with metallic interconnects. In this study, we introduce a versatile circuit that emulates a segment of a dendrite which exhibits gain, introduces delays, and performs integration. We show how sound localisation – a biological example of dendritic computation – is not possible with the existing passive dendrite circuits but can be achieved using this proposed circuit. We also find that dendrites can form bursting neurons. This significant discovery suggests the potential to fabricate neural networks solely comprised of dendrite circuits.

## Introduction

In the pursuit of hosting sophisticated machine learning on battery operated devices, engineers are frequently taking inspiration from biology to replicate nature's efficiency[1]. In neuromorphic engineering artificial synapses and neurons are constructed in a variety of technologies from CMOS electronics[2] to photonics[3] in a bid to build physical neural networks which compute at lower power consumptions. This approach has proved successful, with neuromorphic chips demonstrating 100x lower power consumptions for keyword spotting tasks in comparison to GPU approaches[4], and being used in bioinspired sensors such as the dynamic vision cameras[5] and artificial cochlea[6] which expand neuromorphic principles beyond processing and into the sensing domain.

However, while such circuits are bioinspired in their use of neurons and synapses, today's circuits neglect a key component of the biological neuron – its dendrites.

Dendrites are the tree-like structure that channel inputs from distant synapses to the soma of the neuron where integration and action potential generation will occur. To date, neuromorphic circuits treat dendrites as simple wires, transmitting action potentials from A to B with minimal distortion, but is this a valid approximation?

Dendrites have both passive and active properties. As a passive element they attenuate, broaden[7] and delay[8] action potentials. As an active element they apply gain[9,10], modulate that gain based in local synaptic activity[11], implement a range of integration[12] from sublinear to supralinear and implement coincidence detection. What's more, dendrites have been shown in biology to aid in computation. The clearest example being in the localisation of sound[13], where the delays induced by dendrites cause neurons to be highly sensitive to inputs coming from the left and right ears with specific timing separations. Dendrites are sophisticated computational channels – they are not just a wire.

As a result, neuromorphic engineers are now turning to dendrites to increase the computational power of neuromorphic circuits[14–16]. Circuits replicating biological dendrites can be categorized into two groups. Either they treat the dendrite similar to a communications channel which transforms and distorts the signal[17–19], or, the circuits resemble that of a neuron, producing a voltage spike when ion concentrations satisfy particular conditions[20,21]. We will refer to these two circuits as transmission dendritic circuits and spiking dendritic circuits.

Transmission dendritic circuits are based of the work of Rall et. al. [22] who characterised the transmission properties of biological dendrites. A fundamental finding of this work was that dendrites could be approximated with a Resistor-Capacitor (RC) delay line. The membrane capacitance is modelled with a capacitor, while leakage currents are modelled with a resistor pulling the capacitor to a resting potential or ground. The axial currents (those between neighbouring segments of dendrite) are modelled with a resistor connecting neighbouring RC delay lines.

The RC delay line is naturally suited to being replicated with electronic circuitry. A simple passive circuit can be constructed with resistors and capacitors. However, in integrated CMOS circuits resistors are often replaced with MOSFETs biased in the subthreshold regime [18,23,24]. It is argued that using this specific regime of the MOSFET more accurately represents the migration of ions through the axon which is governed by diffusion because within this regime both electrons and holes also migrate via diffusion. Further, the nonlinearity of the subthreshold regime is argued to be more bio-realistic than the linear conductance of a resistor. In a bid to increase the flexibility of the circuit, and to achieve high resistances within an integrated circuit, the axial resistor connecting two neighbouring sections of dendrite has occasionally been replaced with a switched capacitor [17,25]. However, this does come at the cost of additional clocking circuitry.

Regardless of how the RC delay line is implemented, the key properties of the circuit are its ability to introduce delay to the action potential and to broaden its waveform. These transforms can prove useful in computations such as coincidence detection. However, there are significant limitations to this approach.

The passive nature of the RC delay lane can prove a disadvantage and is also a divergence from biological examples which have been shown to amplify action potentials [9–11]. In a passive circuit, as dendritic branches become longer, or as a single branch divides into many branches, the action potential is attenuated until the point at which it is no longer sufficient to excite the following section of dendrite. This places a limit on the size and complexity of the network and has led to arguments that gain is a requirement to transmit action potentials along longer dendrites such as within motor neurons [26]. To highlight this point, we will discuss in a later section of this paper an application of dendritic computation which requires gain in the dendrites for the computation to be possible. Note that this is also an issue for memristive dendrites.[27] To summarise, if large and complex dendritic networks are to be built, then gain must be possible within the dendritic network.

Introducing gain to the passive RC circuit has previously been done by either adding spiking circuitry to nodes between dendrites [18]

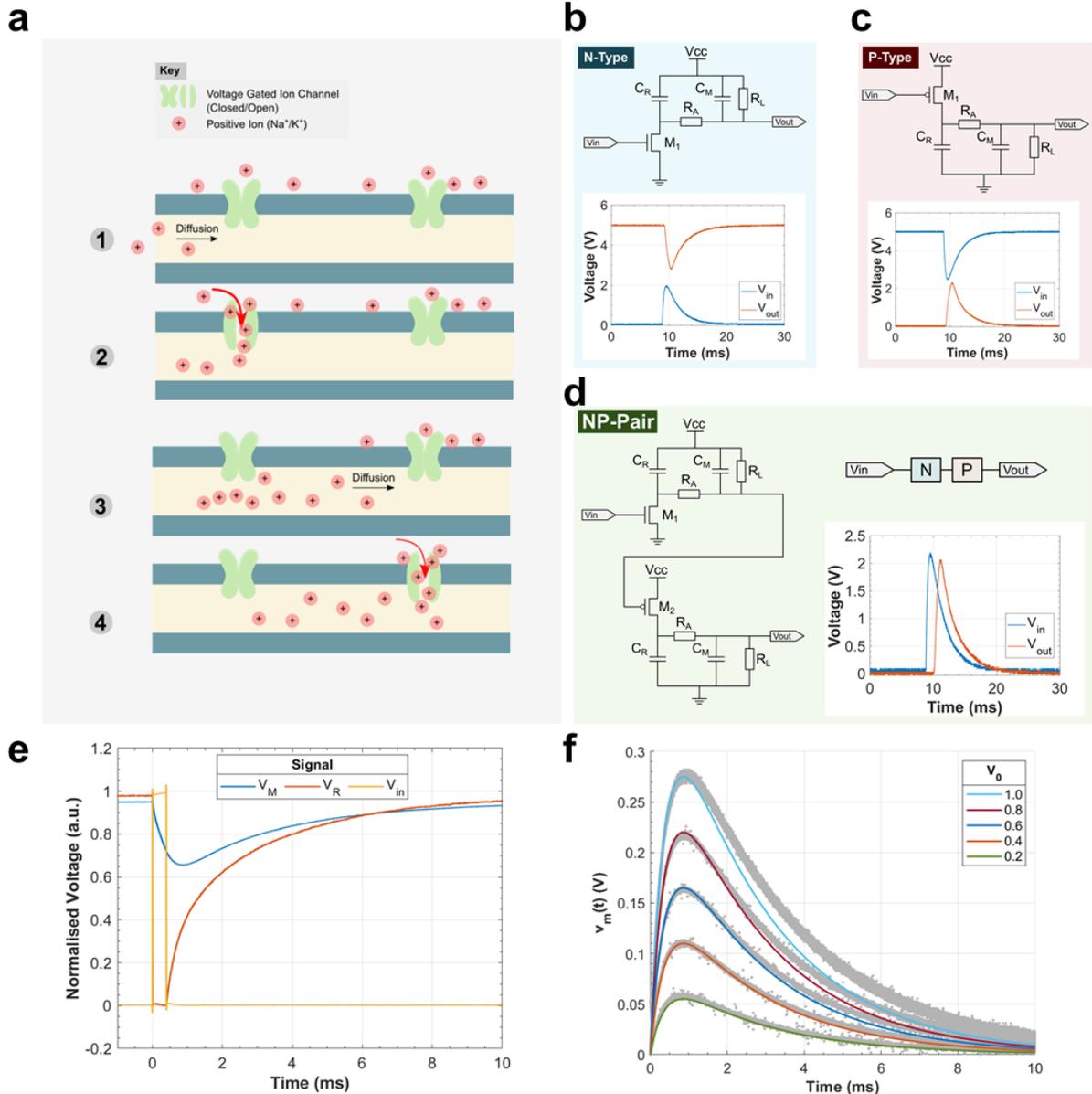

**Fig. 1 | Emulating a segment of dendrite. (a)** The propagation of an action potential within a segment of dendrite is simplified and illustrated in four separate stages. This is the biological process our circuit aims to replicate. **(b)** The proposed circuit diagram of a segment of dendrite constructed using a n-type MOSFET. This is referred to as the n-type variant of our circuit. Inset is a plot of experimental data demonstrating the circuit's response to a spike at its input. **(c)** A second variant of the proposed circuit which instead uses a p-type MOSFET. This is referred to as the p-type variant. Inset is a plot of experimental data demonstrating the circuit's response to a spike at its input. **(d)** The schematic of the np-pair is illustrated alongside experimental data demonstrating the circuit's response to a spike at its input. **(e)** The response of the n-type variant is plotted for a square wave pulse at its input. The voltages are normalised to simplify the comparison of the signal's timings. **(f)** The theoretical response of the resistor-capacitor network is plotted for a range of initial conditions and compared against experimental data. $V_0$ refers to the initial voltage drop across the reservoir capacitor $C_R$. The coloured traces each correspond to the solution of equation 1 for a specific initial condition defined within the legend. The grey points indicate data obtained from a physical implementation of the circuit with each trace being an amalgamation of five separate trials.

or to build dendrites using spiking circuits[21,28]. These circuits generate action potentials with large amplitudes in response to weaker action potentials at their input. They are similar to repeaters used in communication systems. Both of these approaches attempt to address the issue of action potentials decaying as they propagate through the dendritic tree.

Although the active RC delay line circuits that have previously been proposed do indeed introduce gain to the system, it comes at a cost. The spiking circuitry of Farquhar et al.[18] introduces an additional six transistors, two capacitors and seven voltage biases. Equally, in an alternative implementation of active bidirectional dendrites an additional thirty-one transistors are required along with three capacitors and three voltage biases[29]. These circuits do a good job in introducing gain, but the additional components and biases increase the amount of space taken up by the dendritic circuit and the number of voltages which must be distributed throughout the network. Ideally, we wish to minimize both these factors so networks can be made that are dense and small in size without facing issues of routing numerous voltages across the entire network.

In this paper we will introduce an augmented RC delay line which features gain but with a minimal component count of one transistor, two resistors and two capacitors. We will first describe the biological concept it attempts to replicate, then provide an analytical equation describing the circuit's behaviour and demonstrate key dendritic properties the circuit can replicate. We will also introduce an example of dendritic computation in the task of locating the source of an incoming sound. This application will highlight the need for gain within dendritic circuits and demonstrate the versatility of our proposed circuit in it implementing every step of the computation which includes inducing delays, applying gain and carrying out integration. Finally, we will show a unique and novel use of artificial dendrites, to imitate a bursting neuron.

## Circuit Design & Analysis

Our proposed circuit mimics the behaviour of a segment of a biological dendrite. It is intended to be a unit cell which can be chained together in tree like structures to achieve computation. The circuit reproduces the propagation of action potentials by replicating the charging and depolarisation of the dendrite. Fig. 1a illustrates a simplification of this process. First, there is some input charge causing the initial segment of the dendrite to be charged (1). This leads to a change in the voltage difference between the interior and exterior of the dendrite's wall causing voltage gated ion channels to open leading to further depolarisation resulting in an action potential (2). This influx of charge continues to diffuse into the neighbouring section of dendrite, again modifying the electric potential (3). If this change in voltage is significant enough, it will cause the voltage gated ion channels of this neighbouring segment to open leading to further depolarisation (4). This process will repeat along the length of the dendrite assuming the charging of the dendrite is sufficient enough to trigger the ion channels of the adjacent neighbouring segment. It is this process our circuit attempts to replicate.

A key feature of this process is the notion of momentum. Assuming the neighbouring segment is depolarised sufficiently, the voltage gated ion channels remain open and capable of charging the dendrite for some defined time period regardless of whether the initial segment of dendrite is now discharged and no longer enough to keep open the voltage gated ion channels. This means the action potential can continue to propagate along the chain despite the input having died away. This behaviour isn't possible with passive cable models of the dendrite.

Our circuit to replicate this charging behaviour consists of two sections, as shown in Fig. 1b and Fig. 1c. The first section in Fig. 1b comprises an n-type MOSFET combined with a resistor/capacitor network, while the second section in Fig. 1c is similar but uses a p-type MOSFET instead. The two sections are essentially complementary to each other. We will first discuss the n-type variant of the circuit - focussing on how action potentials are generated and propagated - and then return to the p-type variant to discuss why the two are needed.

The input of the circuit is located at the gate of the n-type MOSFET, $M_1$. The voltage at this pin represents the voltage of the preceding segment of dendrite which will go on to trigger the charging of the current segment of dendrite. The output of the circuit (the voltage of the current segment of dendrite) is represented by the voltage across the capacitor $C_M$. We will refer to this as the membrane capacitor because of its symmetry with the membrane wall of the dendrite. The second capacitor, $C_R$, is referred to as the reservoir capacitor because it acts to continue discharging the membrane capacitor even after the input voltage has died away and the transistor has switched off.

The output of this circuit, $V_M$, in response to a single square wave pulse is plotted in Fig. 1e. When the input voltage at the gate of the transistor, $V_{in}$, exceeds its switching threshold, the transistor will switch on, in turn connecting the capacitor at the drain of the transistor, $C_R$, to 0V via the channel of the transistor which has some finite drain-source resistance. This discharges capacitor $C_R$ from VDD towards 0V as indicated by $V_R$. With there now being a voltage difference between the capacitor $C_R$ and the membrane capacitor $C_M$ a current flows between the two capacitors through the axial resistor, $R_A$, in turn causing the membrane capacitor to discharge as shown in Fig. 1e. This causes the output voltage to drop from VDD during the high portion of the pulse.

When the input voltage decays away and falls below the threshold voltage of the transistor, the transistor will switch off causing both the reservoir and membrane capacitors to be recharged to VDD via the leakage resistor, $R_L$, and axial resistor $R_A$. The output voltage of the circuit therefore exhibits a transient similar in shape to an action potential as seen in Fig. 1e.

The shape of this transient is largely defined by the RC network of the circuit. From a circuit analysis detailed in S1 of the Supplementary Materials we determine that the output waveform is defined by equation ( 1 ), where the coefficients $D_+$ and $D_-$ are defined by equations ( 2 ) and ( 3 ) and while the exponents $\lambda_+$ and $\lambda_-$ are defined by equation ( 4 ). $V_0$ denotes the voltage initially across $C_R$ at $t = 0$.

$$v_M(t) = (1 + R_A C_R \lambda_+) D_+ exp^{\lambda_+ t} + (1 + R_L C_M \lambda_-) D_- exp^{\lambda_- t} \quad (1)$$

$$D_+ = \frac{V_0}{1 - \frac{1 + \lambda_+ R_A C_R}{1 + \lambda_- R_L C_M}} \quad (2)$$

$$D_- = -D_+ \frac{1 + \lambda_+ R_A C_R}{1 + \lambda_- R_L C_M} \quad (3)$$

$$\lambda_\pm = \frac{-B \pm \sqrt{B^2 - 4AC}}{2A} \quad (4)$$

Where:

$$A = R_A C_R C_M, \quad B = C_R + C_M + \frac{C_R R_A}{R_L}, \quad C = \frac{1}{R_L}$$

The curves obtained from this expression are plotted in Fig. 1f. Component values are set to generic values: $R_A, R_L = 1k\Omega$, and $C_R, C_M = 1\mu F$. Each curve corresponds to a different initial charging voltage across the reservoir capacitor $C_R$. As can be seen from the curves, an initial pulse of charge across the reservoir capacitor results in a smooth voltage pulse at the output resembling an action potential.

For each case, the data obtained from a physical circuit is plotted alongside the theoretical response. Equation ( 1 ) predicts the circuit's behaviour accurately for lower voltages, $V_0 \leq 0.6V$, however the decay portion of the curve departs from the experimental data for voltages outside of this range. This is most likely due to parasitic resistances having a greater impact at larger voltages. However, also note that any mismatches of component values between the theoretical model and physical circuit will lead to progressively larger errors as time progresses due to the differential nature of the model. Hence, this is also a contributing factor for the increasing error observed over time.

An important characteristic of this circuit, something not possible with passive RC dendrites, is the ability for the membrane voltage to continue charging despite the input voltage no longer being present, a concept we refer to as momentum. In Fig. 1d we can see the output voltage continues to decay despite the input voltage no longer being applied. This is possible because of the reservoir capacitor, $C_R$.

However, while this circuit has produced an output similar to an action potential we need to acknowledge that the output is not yet appropriate to drive the input of another segment of dendrite. The output action potential has a constant offset of VDD and is inverted, that is, while the original action potential had a positive polarity, the output has a negative polarity. To correct this, we need the second half of the circuit, the p-type variant.

The p-type variant of the circuit operates in the similar manner as the n-type, however it is the complement of the circuit. The input of the circuit is again at the gate of the transistor, although the transistor is now a p-type MOSFET. The transistor is therefore switched on when the gate drops below VDD by more than the threshold voltage of transistor. The output of the circuit is again the voltage across the membrane capacitor $C_M$ and will be the inverted form of the input. The fact both the n-type and p-type circuits invert the signals at their input, means that by chaining the two together we can obtain an output that is of the same polarity as the input. We refer to this combination as an np-pair.

## Circuit Characterisation

In this section we will demonstrate the variety of behaviours the circuit exhibits and relate this to the transmission properties observed in biological dendrites. We will address the gain and attenuation of the circuit, the delay it introduces and its integration properties. These are key properties which can give arise to computation and are therefore crucial for the circuit to exhibit.

Note that the characterisation of any dendritic circuit suffers from a common challenge, which is the inherent nonlinearity of the circuit. Like their biology counterparts, artificial dendrites are highly nonlinear; their transmission properties such as gain and delay will vary depending on the amplitude of the input signal as well its shape. It's therefore not possible to provide a comprehensive characterisation of the circuit that will apply to any input. Instead, we will present generalised examples of the different dendritic behaviours the circuit is capable of replicating and then provide a more comprehensive analysis which is restricted to a limited range of input signals and component values.

### General Response

To demonstrate the general behaviour of each circuit we have plotted the responses of the n-type, the p-type and the np-pair to a single action potential like input in Fig. 1b, Fig. 1c and Fig. 1d respectively. The component values were fixed and identical for each variant of the circuit: $R_A = R_L = 1k\Omega$ and $C_R = C_M = 1\mu F$. The input wave form is generated by a signal generator and applied to the gate of the input transistor. The input waveform is generated from equation ( 1 ) to mimic chaining of multiple dendrite circuits.

From these responses we can identify a number of basic dendritic properties which we will explore in more detail later. For example, we observe a delay in the action potential on the order of milliseconds for each version of the circuit. We also observe a gain > 1 in the n-type version of the circuit, Fig. 1b, which is crucial in being able to construct larger dendritic trees without suffering the effects of attenuation which would limit the possible size of networks.

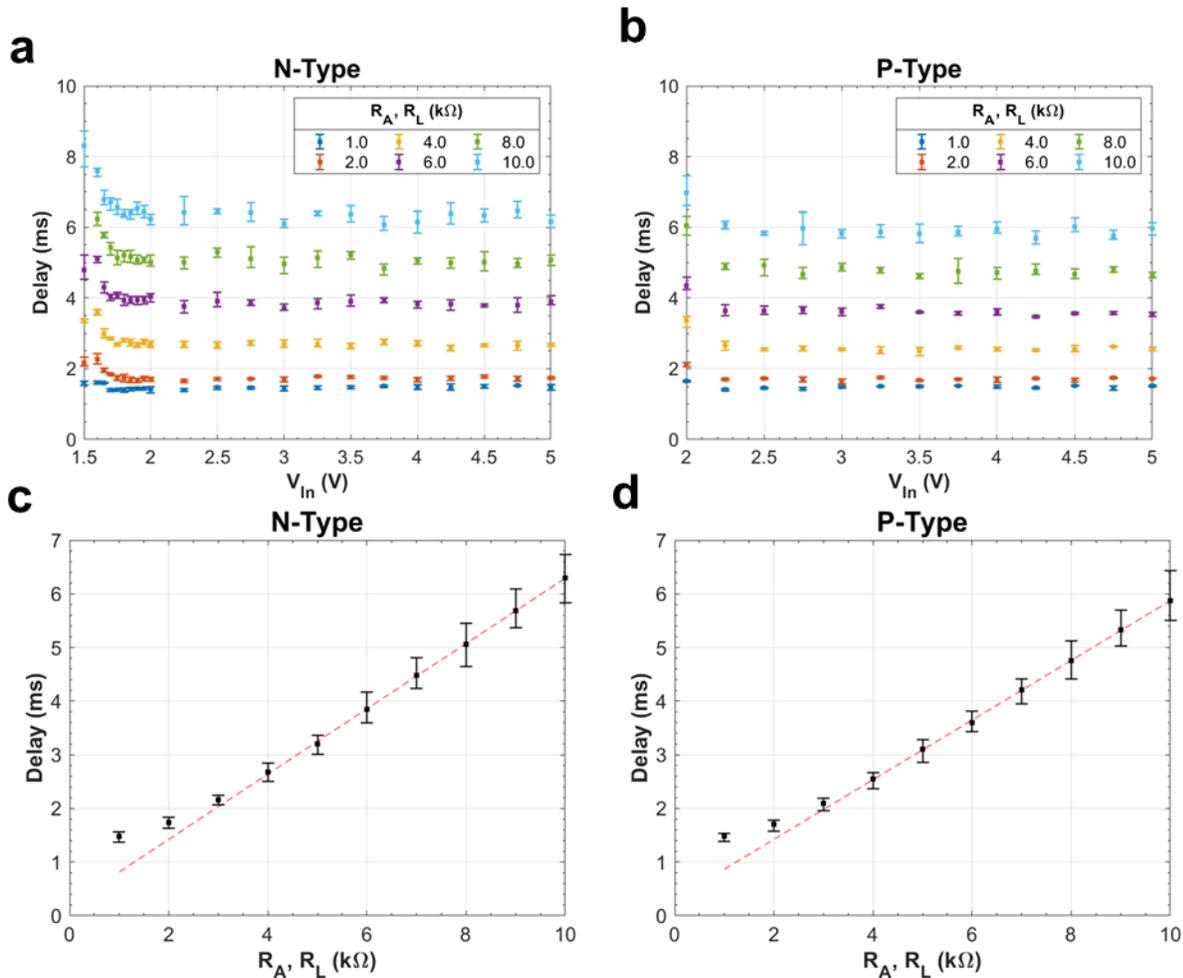

**Fig. 2 | Delay characteristics of the dendrite circuits. (a, b)** The delay induced by the n-type and p-type circuits is plotted for a range of input amplitudes and for increasing axial and leakage resistances where $R_A = R_L$. The input was a square wave pulse detailed in the main body. Error bars indicate the range of three separate trials, while data points indicate the mean average delay. **(c, d)** The delay within the saturation region of the n-type and p-type circuits are plotted as a function of the axial and leakage resistances where $R_A = R_L$. The saturation region is defined by input pulses having an amplitude $> 2V$. Each datapoint for a given resistance value is the average of 36 trials, with input signals exhibiting a range of input amplitudes from $2V$ to $5V$ in increments of $0.25V$. There are three trials for each input amplitude. The error bars indicate the range of the trials.

Not only does the circuit replicate the action potential waveform at its input but the circuit can also be used to generate similar waveforms from square wave pulses as predicted by our previous circuit analysis. For example, in Fig. 1e we show the circuit's response to a square wave pulse at its input which generates an action potential like waveform at the output. This could prove useful when interfacing between digital systems and more bio-realistic circuits.

We will now investigate each of these key transmission properties in closer detail.

**Delay**

Dendrites introduce delays to the propagation of action potentials travelling along their branches. These delays have been shown to be useful in computation within dendritic trees [13,26,30]. Replicating these delays is therefore an important task of artificial dendrite circuits.

The circuit presented in this paper is capable of inducing delays to the action potential applied to its input. The induced delay is defined by the time difference between the peak of the input and output signals. This is governed by the time constants of the RC network formed by $R_A$, $R_L$, $C_R$ and $C_M$. However, it is also affected by the degree to which the reservoir capacitor is discharged and therefore also depends on the amplitude of the input pulse resulting in a nonlinear circuit. Due to the large number of variables which affect the delay properties we cannot provide a comprehensive documentation of the circuit's behaviour. Instead, we present a family of curves where capacitances are fixed, $C_R = C_M = 1F$, the resistances $R_A$ and $R_L$ are increased from $1k\Omega$ to $10k\Omega$ and the magnitude of the input pulse is swept from $1.5V$ to $5V$. Delays are not reported for pulses with amplitudes $< 1.5V$ for the n-type and $< 2V$ for the p-type because the amplitude of the output pulse is too small to accurately measure the delay.

As expected, the delay of the circuit increases with larger resistances. In Fig. 2a and Fig. 2b the delays of both the n-type and p-type circuits are plotted for input pulses with increasing amplitudes and resistances. Initially, the delays of inputs with smaller magnitudes ($< 2V$ for the n-type and $< 2.5V$ for the p-type) exhibit a voltage dependency. This is caused by the transistor only partially discharging the reservoir capacitor and the degree to which it is discharged depends on the magnitude of the input pulse. However, beyond this range, the reservoir capacitor is fully discharged and the delay becomes constant. A circuit can therefore be designed for a constant delay so long as the input voltage exceeds this range. We will refer to this region as the saturation region.

Within this saturation region, the delay is approximately linearly dependent on the resistance values of $R_A$ and $R_L$. Note, that this is in the case that $R_A = R_L$. In Fig. 2c and Fig. 2d the delays of the n-type and p-type circuits are plotted for a range of resistance values. To obtain the delays for a specific resistance value, we combine the delays measured for all inputs with amplitudes $> 2V$, thereby taking into account the variance that occurs within the saturation region. The linear dependence holds for larger resistances, $> 4k\Omega$, however this breaks down at lower resistances. With this in mind, larger resistances are preferred when designing a circuit to exhibit a particular delay because the linearity simplifies component selection.

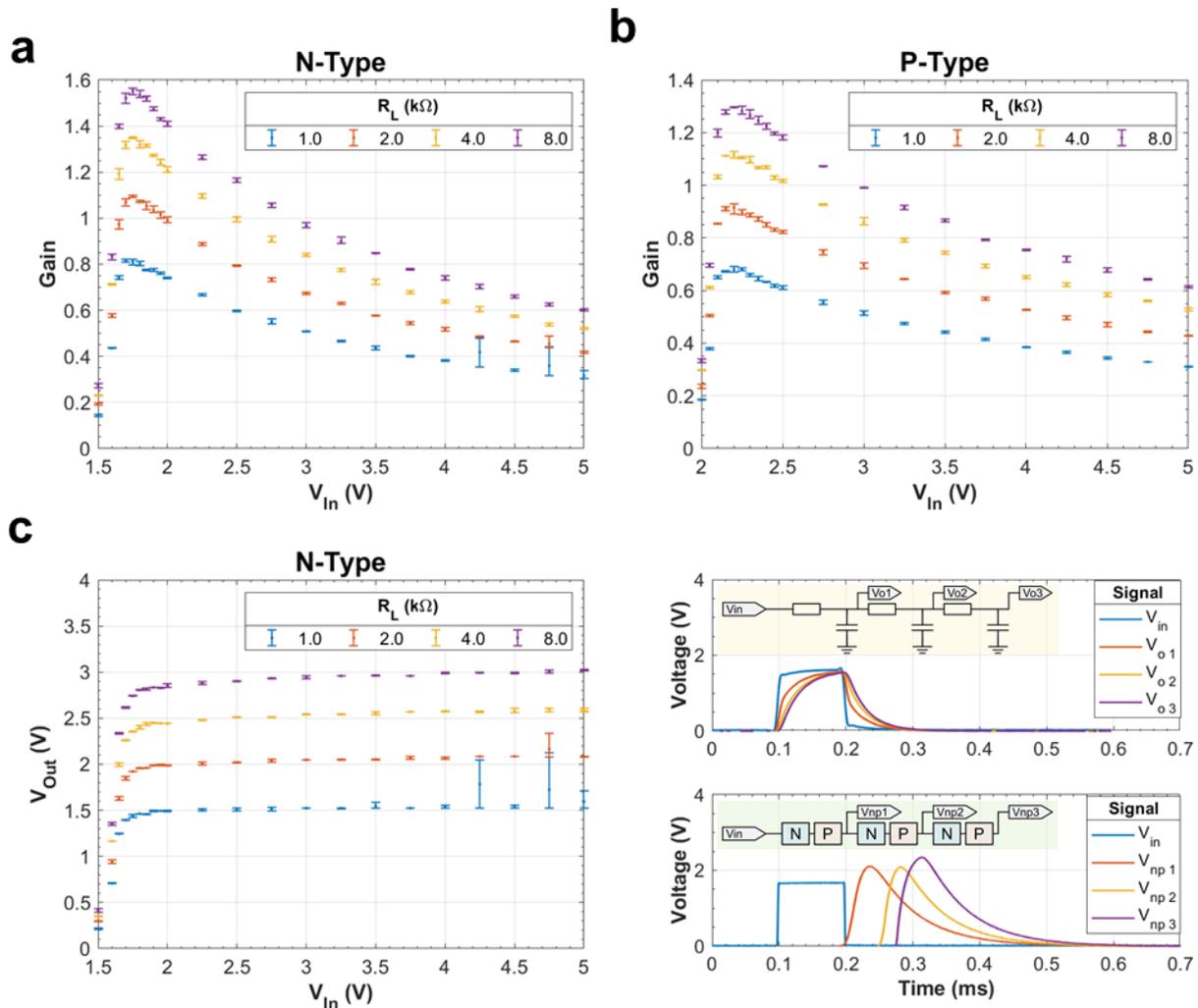

**Fig. 3 | Gain characteristics of the dendrite circuits. (a, b)** The gains of the n-type and p-type circuits are plotted as a function of input amplitude and increasing leakage resistance $R_L$, while the axial resistance is fixed to $R_A = 2k\Omega$. **(c)** The amplitude of the output action potential from the n-type circuit is plotted as a function of the input pulse's amplitude and an increasing leakage resistance $R_L$. The output is found to saturate, which is the cause of the decreasing gain within the saturation region. **(d)** The propagation of an action potential travelling along a chain of passive dendrites is compared to our own active implementation. It highlights the need for gain if larger networks are to be constructed.

**Gain**

The ability of a dendrite to apply a gain greater than 1 to an action potential is crucial for the propagation of action potentials across longer dendritic branches. An example of the circuit amplifying a signal was shown in Fig. 3a where an input square wave pulse was applied to the input of the circuit. Note that the circuit always has a negative gain. To obtain a noninverting dendrite segment two copies of the circuit are placed in series for example an n-type followed by a p-type, forming an np-pair.

However, as was the case for the circuit's delay, the gain is not linear but depends on the amplitude of the input signal. The circuit's gain as function of maximum input amplitude is plotted in Fig. 3a and Fig. 3b for the n-type and p-type circuits each with different values of $R_l$. When the input signal is small, the transistor remains highly resistive, resulting in no signal at the output and a negligible gain. As the amplitude of the input signal increases beyond 1.5V for the n-type and $2.0V$ of the p-type, the gain increases rapidly. This is the result of the drain-source resistance of the transistor reducing with an increased input voltage as the transistor begins to switch on. Above a threshold voltage of approximately 1.7V for the n-type and $2.2V$ for the p-type, the output of the circuit saturates and the amplitude of the output waveform can no longer increase with increases in the input amplitude – resulting in an apparent reduction in gain.

The maximum gain of the circuit is limited by the potential divider formed by $R_a$ and $R_l$ and the degree to which the reservoir capacitor is discharged. In Fig. 3a and Fig. 3b, $R_a$ is fixed at $R_a = 2k$ while $R_l$ is swept from $1k\Omega$ to $8k\Omega$. The voltage supply is fixed at $5V$. It's clear the gain of the circuit is heavily voltage dependent, initially the gain increases rapidly with increasing voltage until $V_{in} = 2V$. Beyond this, the output voltage saturates as shown in Fig. 3c, which is the cause of the decreasing gain at higher input amplitudes.

The ability of the circuit to amplify action potentials has advantages over the passive RC cable theory circuits used in existing studies. One advantage is the possibility to chain many circuits in series without risk of action potentials being attenuated to the point of being negligible. In Fig. 3d we contrast the chaining together of a passive RC dendrite and our active dendrites. For the passive dendrites the action potential's amplitude remains constant with a slight attenuation as it progresses along the chain, whereas for the active dendrites the action potential is amplified as it travels along the chain. This enables us to build larger networks which would not have been feasible with passive RC dendrites.

Note that our circuit is not the only active implementation of dendrites. Other studies have successfully developed active dendrites emulating spiking behaviours[18,20,21]. These circuits are able to introduce gain and have been used in a variety of applications including as repeaters to amplify dendrites within passive RC networks. However, such circuits struggle with requiring a larger number of transistors and biases than the circuit we propose. This has the disadvantage of the dendrite requiring more physical space, although it also allows for the dendrite to be more tuneable. In contrast, our circuits behaviours are largely fixed at fabrication. Depending on which circuit is more appropriate for a given application is based on which of these two characteristics are priorities: a smaller physical footprint, or a more plastic and malleable system.

**Integration**

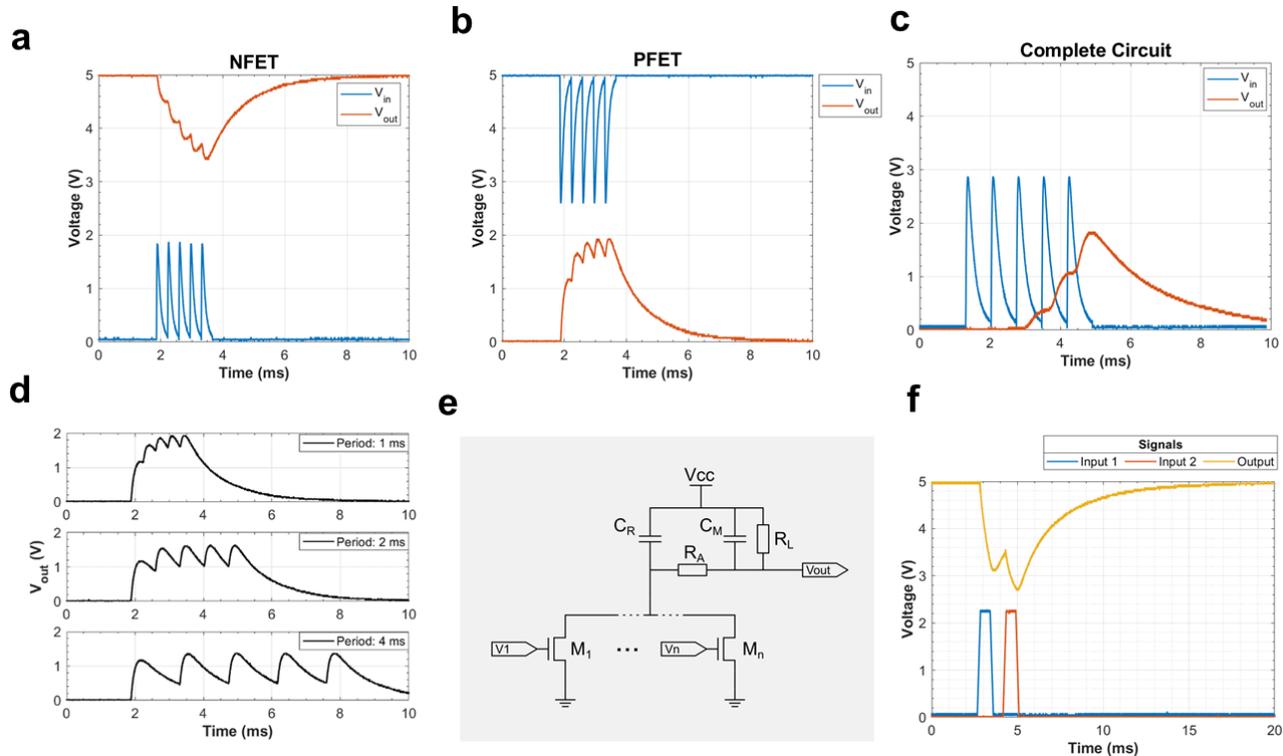

**Fig. 4 | Integrating with dendrites.** The circuit's output to four sequential spikes is plotted for the **(a)** n-type, **(b)** p-type and **(c)** np-pair variants of the circuit. In all circuits the components value are fixed: $C_R = C_M = 1\mu F$ and $R_A = R_L = 1k\Omega$. **(d)** The leaky nature of the integration is demonstrated for the p-type version of the circuit by increasing the period between subsequent pulses. Clear integration is observed with a period of $1ms$ while integration becomes negligible when pulses are separated by more than $4ms$. **(e)** The circuit diagram detailing how spatial integration of multiple inputs onto a single dendritic branch is achieved by attaching multiple transistors in parallel to the reservoir capacitor $C_R$. **(f)** Two square wave pulses with a width of 2ms are applied to an n-type circuit with two separate inputs. The output of the circuit demonstrates the integration of these two inputs with the second peak higher than the first. The integration is sublinear but shows the possibility of integrating multiple incoming dendritic branches into a single segment.

The circuit also replicates the integration properties of the dendrite both temporally and spatially. In Fig. 4 we demonstrate the circuit's ability to integrate temporally. A signal generator is used to apply a series of action potentials to the input of the n-type, p-type and np-pair variants. In all cases a clear leaky integration is observed for each spike train. The leakiness of this integration is best demonstrated by comparing the circuit's response to spikes with varying periods. In Fig. 4d the p-type circuit's responses to spike trains with periods of 1, 2 and 4ms are plotted. An accumulation is observed in the spike train of 1ms whereas minimal accumulation occurs for spikes with a period of 4ms, due the leakage properties of the circuit's integration.

A segment of dendrite is also required to receive and integrate inputs from multiple different inputs – referred to as spatial integration. This achieved by adding additional transistors to the reservoir capacitor as shown illustrated in Fig. 6e. To demonstrate this concept the p-type variant of the circuit is constructed with two input p-type MOSFETs. The input signals and corresponding output of the circuit are shown in Fig. 6f with clear accumulative integration occurring. The first pulse generates the action potential like curve that is typical of the circuit. The second pulse produces the same pulse however with a greater amplitude due to the integration occurring from the previous pulse. The type of integration occurring is sublinear, that is the total is less than the sum of the two inputs. This is most clear when the two pulses overlap exactly and the amplitude of the input is not twice that of a single pulse but smaller than expected. In biological samples a range of integration is observed from sublinear to supra-linear.[12]

## Enabling Sound Localisation

In this section we will demonstrate the advantages of providing gain within the dendritic circuitry by showcasing one application of dendritic computation which is not possible without gain. We will also demonstrate the circuit's versatility showcasing its ability to induce delays, apply gain and carry out integration all within the same circuit.

One example of dendritic computation is coincidence detection within the auditory nerves[13]. Neurons have been documented detecting the timing of inputs from both ears. This allows the subject to localize a sound based on the differences in the time of arrival at the left and right ears.

For a neuron to respond maximally to inputs separated by a specific timing interval, dendrites are needed to induce delays to the inputs. These delays act to bring the two inputs back in time with one another which occurs if the dendrite's delay matches the timing separation of the inputs as illustrated in Fig. 6a. By producing pairs of dendrites with a range of delays and connecting these to neurons, we obtain a bank of neurons each of which is most active in response to inputs separated by particular timing intervals. It is estimated that this concept allows mammals and birds to detect timing intervals on the order of 10-100$\mu s$ [31–34]. Further, it has been shown that by employing dendrites to introduce delays to the inputs of a neural network the network can be reduced in size, by up to a half, yet still maintain the same accuracy as a larger network without input dendrites [30].

The passive RC delay line is capable of producing delays, however, due to the circuit's attenuation its implementation becomes impractical. With the RC circuit, as the action potential is delayed to greater degree it also broadened and attenuated to a greater extent. This makes comparing dendritic branches of different delays a challenge. Dendritic branches with little delay will have larger amplitudes at their output than the branches with longer delays. This is an issue because it should be the dendrite who's delay matches the timing separation of the input that has the largest output so as to trigger its attached neuron more so than its neighbouring neurons.

In Fig. 6b and c, we demonstrate this problem. Multiple pairs of dendrites have been constructed with RC delay lines each of which induce a different delay to one of the inputs by increasing the axial resistance, Fig. 6b. At each pair of dendrites, the output from two dendritic branches are added together to replicate the integration properties of the neuron. The maximum value of this summation is taken as the output strength of the dendritic branch. We have then characterized each dendritic branch for inputs with a range of timing intervals, Fig. 6c.

Each pair of dendrites has a peaked response whose maximal values corresponds to the delay induced by one of the two branches. However, the peak of the dendrites with greater delays do not exceed their neighbouring dendrites. To correct for this, the dendritic circuit must be able to apply gain. The circuit we have proposed in this paper is able to apply both delays and gains. For this reason, it overcomes the challenges faced by the passive RC delay line dendrites.

Our circuit to replicate the timing interval detection is illustrated in Fig. 6d. It consists of two n-type dendrite circuits whose outputs are then integrated by a single p-type dendrite circuit. The two n-type circuits are tuned to apply the desired delay and gain to the action potential. The final p-type circuit detects how aligned the two action potentials are at the output of the n-type dendrites via the temporal and spatial integration properties we demonstrated earlier. The amplitude of the output action potential of the p-type dendrite will be maximal when the two inputs are in phase. This also highlights the versatility of the circuit's design. We can use the same circuit to carry out three different tasks: apply gain, introduce a delay and detect the coincidence of inputs.

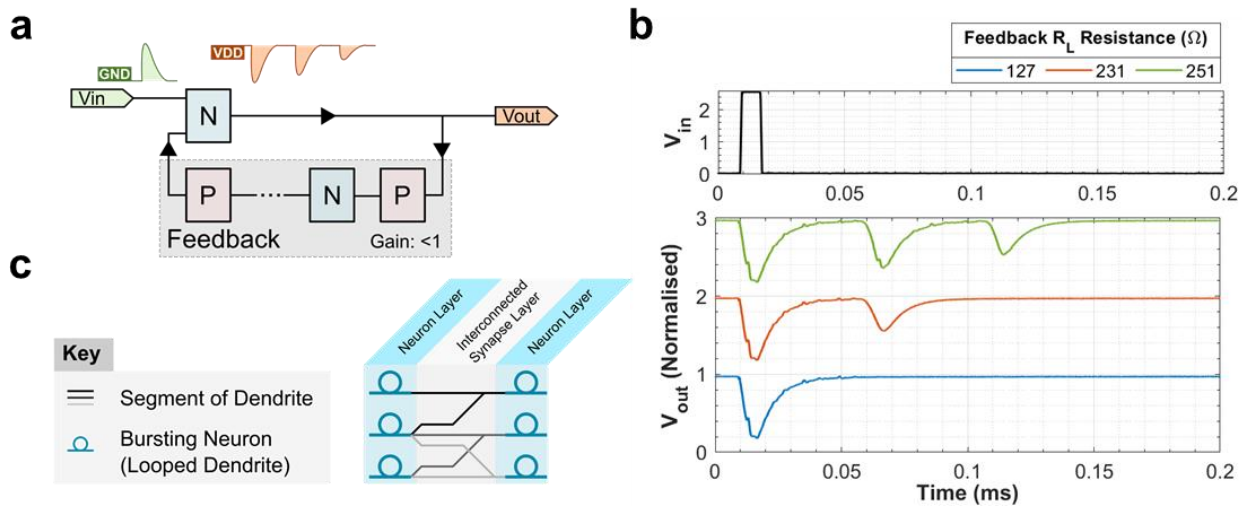

**Fig. 5 | Imitating neurons with looped dendrites. (a)** The schematic of a bursting neuron implemented with dendrite circuits. The gain of the feedback loop is tuned to modulate the number of spikes within a burst. **(b)** Experimental data from a bursting neuron implemented with looped dendrites. The number of output spikes in response to a single input pulse is modulated by adjusting the gain of the feedback loop through the leakage resistance of the final p-type dendrite. The voltages have been normalised and offset for clarity. **(c)** An illustration of a dendritic neural network. Chains of dendrites connect subsequent neuron layers, while looped dendrites imitate bursting neurons.

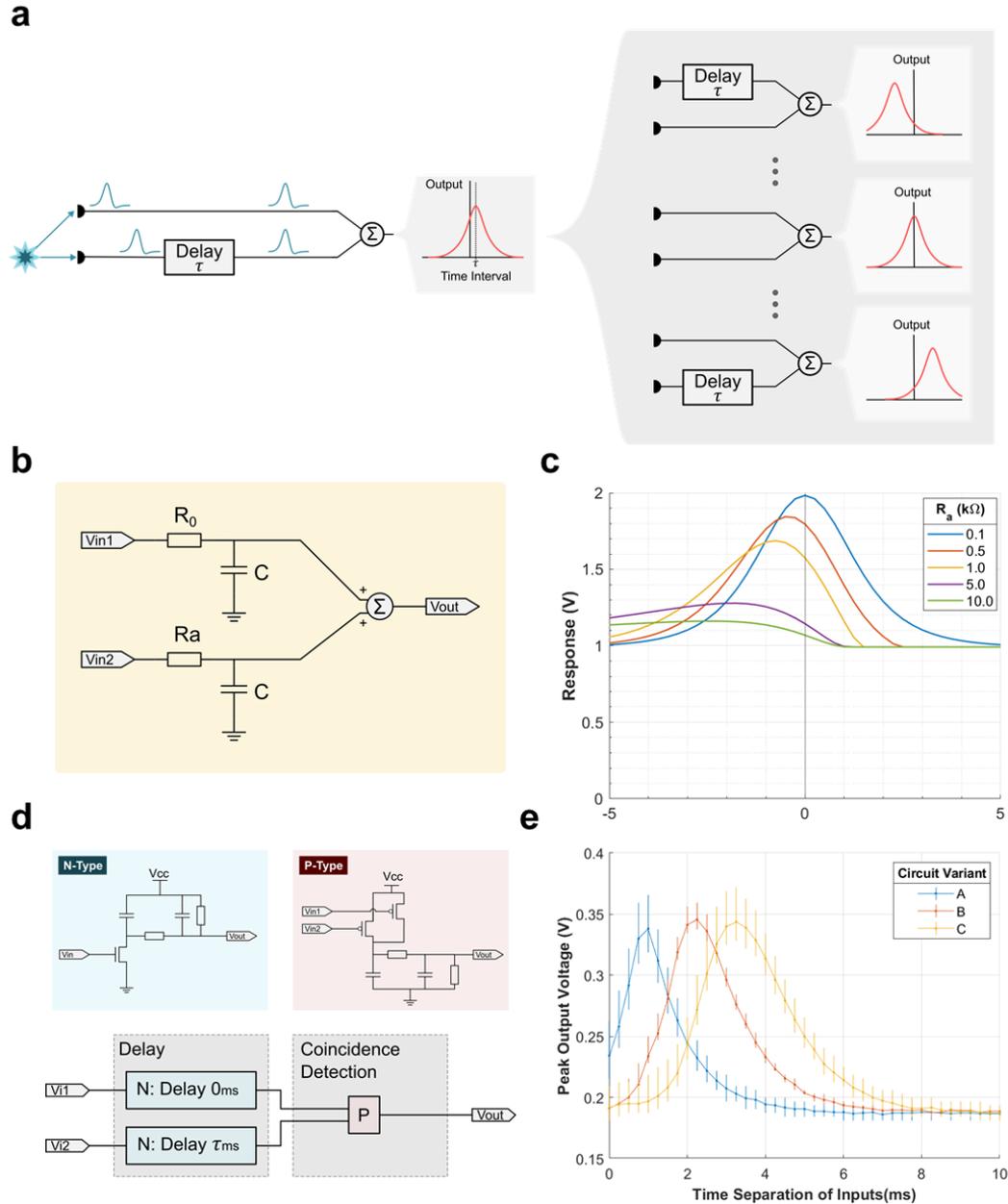

**Fig. 6 | Enabling sound localisation via the introduction of gain. (a)** An illustration of a dendritic tree designed to detect the timing separation between two input signals. The two inputs travel on separate branches, each of which introduce a different delay. If the difference in delay of the two branches matches the timing separation of the inputs the pulses are brought inline with each other and will produce a maximal output when integrated by a leaky integrator. By modulating the difference in delays of the two branches, the time difference which the dendritic tree is most resonant to is shifted. **(b)** The schematic diagram of the passive RC circuit attempting to replicate the task. **(c)** The simulated response curves of five different variants of the passive RC circuit each with an increasing resistance in one of the two branches, $R_A$. The peak of the curve shifts as desired; however, it is also attenuated for larger delays which is not ideal. **(d)** The proposed block and circuit diagram of the sound localisation dendritic tree using the n-type and p-type dendrite circuits. The input signals are applied to two n-type dendrites designed to induce delays. The outputs of these circuits are integrated by a p-type dendrite designed to induce minimal delay and detect the coincidence of the two inputs. **(e)** The response curves for physical copies of the n-type and p-type detector with three variants each designed to detect progressively larger timing separations. As expected the peak shifts and the gain of the n-type circuits allow longer delays to avoid attenuation and maintain the same amplitude as shorter delays. Each data point corresponds the mean average of five trials with the vertical bars indicating the range of the five trials.

We have constructed three variants of the circuit in Fig. 6d, each designed to detect a particular timing interval. The tuning of the circuit can be achieved by varying the capacitor and resistance values. In this study we fixed all capacitor values to $1\mu F$ and tuned the resistance values to achieve the desired gain and delay. The final component values are detailed in Table 1 within the Supplementary Materials.

The response of each dendritic tree is summarised in Fig. 6e with the y-axis indicating the maximum voltage at the output of the p-type dendrite. Each circuit features a peaked response to the input's timing interval. Crucially, the peak of each circuit's response, that is the timing interval it is designed to detect, is larger than the output for all the other circuits at that specific timing interval. As discussed earlier, this is crucial to ensuring the attached neuron is the most active and so would correctly indicate the inputs were separated by the corresponding interval.

The response curves of the circuits proposed in this paper therefore contrast significantly with those produced by passive RC circuits used in previous implementations of dendrites [23,25] as shown in Fig. 6c. The response curves of the passive RC circuits progressively attenuate as the circuit is designed to induce larger delays making it more of a challenge to identify which of the dendritic trees corresponds to the current timing separation of the inputs. In contrast, the response curves of our circuits do not attenuate with larger delays, making it clearer which of the dendritic trees has a delay most similar to the input's timing separation.

Introducing gain to the dendritic circuit has therefore made it possible to replicate the sound localisation within the auditory nerves of mammals and birds. Whilst sound localisation is one application of this circuit, the computation can also be generalised. For example, the circuit is essentially carrying out a transform on the incoming data; from time encoding to a spatial encoding whereby the neuron connected to each dendrite corresponds to a particular timing separation. This representation of the data is better suited to being analysed with a neural network. The circuit could therefore play a role in the pre-processing of sensor data prior to being analysed by a neural network.

## Neuron Imitation via Looped Dendrites

A surprising and as of yet undocumented use of dendrites is to imitate bursting neurons. We found that by constructing a loop of dendrite segments bursting behaviour arises. When a single pulse is applied to the input of this circuit, the output generates a train of spikes at its output. The circuit exhibits a threshold, as is required of a neuron and can be designed to output multiple spikes thereby imitating bursting behaviour. The number of spikes within a burst can be modulated by varying either the amplitude of the input pulse, or the gain of the feedback loop as shown in Fig. 5a.

This bursting behaviour is result of the feedback introduced by connecting the output of the dendrite to its input. Crucially, this feedback applies both delay and gain to the feedback signal to ensure stable spiking conditions. This delay and gain is applied using the same dendrite circuitry, highlighting the versatility of the circuit. The delay must be large enough to ensure the input and feedback spikes do not overlap. If not, the output will latch at the supply rails due to the positive feedback that occurs. Equally, the gain of the feedback look should be < 1, again to ensure the spike train attenuates with each spike and does not cause the neuron to saturate to the supply rails. To modulate the number of spikes within a burst, the gain of the feedback is adjusted. This is done by modulating the gain of the final dendrite circuit with the leakage resistor $R_L$ as shown in Fig. 5b.

Loops of dendrites can therefore be used to imitate spiking neurons. This suggests that a more complex dendritic network could introduce neuron-like behaviours within its branches potentially increasing the sophistication of the computation that dendrites can implement. Equally, we have considered only a single loop, but dendritic branches could have multiple loops within its structure allowing for neurons within neurons. We could even consider neural networks which are entirely constructed from the same dendritic circuit whereby dendrites form the interconnects and loops of dendrites form neuron layers, such a network is illustrated in Fig. 5c.

## Conclusion

We have presented a versatile circuit to replicate a segment of a dendritic branch. The impact of this circuit lies in its ability to replicate a range of dendritic behaviours such as: inducing delays, carrying out integration, and most importantly, introducing gain. All of this is achieved within a single circuit.

The ability to apply gain enables both larger and more dense networks, which are not possible with the existing passive dendrite circuits, and enables a key demonstration of dendritic computation – sound localisation. This demonstration also showcased the versatility of our proposed circuit, with it being used to both induce delays as well as integrate multiple inputs to implement coincidence detection.

Finally, we documented the first use of dendrites to imitate neurons. Using only dendrite circuits, we produced a bursting neuron with tuneable properties. Based on this we propose a neural network architecture built entirely from dendrite circuits, with strings of dendrites used to process and distribute action potentials between layers of neurons built from looped dendrites.

With neuromorphic engineers now expanding beyond neuron and synapses architectures to explore the potential of artificial dendrites, this circuit is a timely contribution and a useful tool for testing novel concepts.

## Methods

### Circuit Characterisation

The experimental data obtained in figures 1b-f and figure 4 where obtained from circuits constructed with ROHM SP8M3HZGTB dual MOSFET transistors. Capacitors and resistors had the following values: $C_R = C_M = 1\mu F$ and $R_L = R_A = 1k\Omega$. The circuit was constructed on a bespoke printed circuit board. Input signals were sourced from a Rigol DG1062Z function generator. Output signals were captured with a Rigol DS4024 oscilloscope. Data obtained from the circuits underwent no post processing.

The experimental data obtained for figures 2 and 3 was obtained from circuits constructed with ROHM SP8M3HZGTB dual MOSFET transistors. Capacitances were fixed at $C_R = C_M = 1\mu F$ while resistances were varied as indicated in the figure legends.

The experimental data obtained for figure 3d used circuits constructed with Fairchild 2N7000 n-channel MOSFETs and Microchip TP2104 p-channel MOSFETs. All capacitors were fixed to $C_R = C_M = 22nF$ and all resistances were fixed to $R_L = R_A = 220\Omega$. Input signals were sourced from a Rigol DG1062Z function generator. Output signals were captured with a Rohde and Schwarz RTB2004.

### Sound Localisation

The simulation of the RC delay line was carried out in MATLAB Simulink.

The experimental data within figure 5e was obtained from circuits using ROHM SP8M3HZGTB dual MOSFET transistors. The component values for each variant of the circuit are summarised in Table 1 of the Supplementary Materials. Resistances were tuned using a BOURNS 3266w-1-102LF trimmer in series with a fixed 0805 surface mount resistor.

The inputs to the circuit are square wave pulses with an amplitude of 1.7 V and a width of 2ms sourced from a Rigol DG1062Z function generator. The timing separation of the inputs was swept from 0ms to 10ms in 0.25ms intervals.

### Neuron Imitation

The bursting neuron was constructed using Fairchild 2N7000 n-channel MOSFETs and Microchip TP2104 p-channel MOSFETs. The capacitor and resistance component values are detailed in Table 2 within the Supplementary Materials. Input signals were sourced from a Rigol DG1062Z function generator. Output signals were captured using a Rohde and Schwarz RTB2004 oscilloscope.

# Supplementary Materials

## S1. Circuit Analysis

We will now derive analytical equations which describe the circuit's behaviour. This will aid in determining how particular components affect the circuit's response to a step potential as well as identify component values required for a certain response.

The goal of this analysis is to determine how the output voltage varies over time given the reservoir capacitor has been charged to some voltage by the input transistor. To do this we will derive a differential equation describing the voltage across the reservoir capacitor, $v_R$, solve this equation and finally use this solution to determine the voltage across the membrane capacitor, $v_M$. The following work focusses on the p-type variant of the circuit shown in Figure 1c however a similar analysis applies to the n-type variant.

We begin with a simplification of the circuit analysing only the passive components of the circuit. We assumed the reservoir capacitor is charged to an initial voltage $V_0$ by a pulse at the input transistor, equation ( 1 ), while the output voltage across the membrane capacitor and leak resistor begins at $0V$, equation ( 2 ). We have the following initial conditions:

$$v_R(t = 0) = V_0 \qquad (1)$$

$$v_M(t = 0) = 0 \qquad (2)$$

To begin, we obtain equations describing how the voltage at the output varies with time. The voltage at this node within the circuit is defined by two equations.

First, the output voltage can be determined form the voltage drop across the axial resistor, $R_A$, resulting in an equation expressing the output voltage as a function of the input voltage and axial currents, $I_A$.

$$v_M = v_R - R_A I_A \qquad (3)$$

A second equation that defines $v_M$ is determined by applying Kirchhoff's law at the node of the output voltage. The current injected into the node is equal to the current through the axial resistor, $I_A$, and the total current drawn out of the node is the sum of the current through the leak resistor, $I_L$, and the membrane capacitor $C_M$. This analysis results equation ( 4 ), which relates the output voltage and the current through the axial resistor.

$$I_A = v_M R_L + C_M \frac{dv_M}{dt} \qquad (4)$$

We now derive the differential equation describing the voltage across the reservoir capacitor, $v_R$. To start, we consider the differential equation describing the voltage across the reservoir capacitor itself. The current flowing out of the reservoir capacitor, $C_R \frac{dv_R}{dt}$, is equal to the current through the axial resistor, $I_A$, although opposite in sign, equation ( 5 ).

$$I_A = -C_R \frac{dv_R}{dt} \qquad (5)$$

To derive a differential equation describing $v_R$, we must obtain an expression for the axial current solely in terms of $v_M$. From our previous analysis, the axial current can be expressed as a function of the output voltage, equation ( 4 ), resulting in the following equation:

$$-C_R \frac{dv_R}{dt} = \frac{v_M}{R_L} + C_M \frac{dv_M}{dt} \qquad (6)$$

Furthermore, the output voltage $v_M$ and its derivative, can be expressed in terms of the input voltage $v_R$ using equation ( 3 ).

$$v_M = v_R + R_A C_R \frac{dv_R}{dt} \qquad (7)$$

$$\frac{dv_M}{dt} = \frac{dv_R}{dt} + R_A C_R \frac{d^2 v_R}{dt^2} \qquad (8)$$

Substituting equations ( 7 ) and ( 8 ) into equation ( 6 ) gives the following differential equation describing the input voltage $v_R$.

$$(R_A C_R C_M)\frac{d^2 v_R}{dt^2} + \left(C_R + C_M + \frac{C_R R_A}{R_L}\right)\frac{dv_R}{dt} + \frac{1}{R_L}v_R = 0 \qquad (\,9\,)$$

This differential equation is solved using the characteristic equation. The characteristic equation is given by:

$$A\lambda^2 + B\lambda + C = 0 \qquad (\,10\,)$$

Where:

$$A = R_A C_R C_M$$
$$B = C_R + C_M + \frac{C_R R_A}{R_L}$$
$$C = \frac{1}{R_L}$$

The characteristic equation technique involves finding the two solutions to equation ( 10 ): $\lambda_+$ and $\lambda_-$ using the quadratic equation. These solutions correspond to two different expressions which satisfy the original differential equation, with the form $v_R(t) = D_\pm \, exp(\lambda_\pm t)$, where $D$ is a coefficient determined from the initial conditions. The general solution is obtained by linearly summing these two solutions, resulting in equation ( 11 ).

$$v_R(t) = D_+ exp^{\lambda_+ t} + D_- exp^{\lambda_- t} \qquad (\,11\,)$$

Where:

$$\lambda_\pm = \frac{-B \pm \sqrt{B^2 - 4AC}}{2A}$$

The coefficients, $D_+$ and $D_-$, are obtained from the initial conditions defined by equations ( 1 ) and ( 2 ). We will use these conditions to derive two simultaneous equations, which are then solved to obtain the coefficient's values.

From the initial condition: $v_R(t = 0) = V_0$, we obtain the first simultaneous equation.

$$D_+ + D_- = V_0 \qquad (\,12\,)$$

From the second initial condition: $v_M(t = 0) = 0$ we obtain the remaining simultaneous equation by substituting equation ( 11 ) into equation ( 7 ) and applying the initial condition.

$$v_M(t = 0) = D_+(1 + \lambda_+ R_A C_R) + D_-(1 + \lambda_- R_L C_M) = 0 \qquad (\,13\,)$$

We will now solve these simultaneous equations to obtain the values of the coefficients. Equation ( 13 ) is rearranged to obtain an expression of $D_-$ in terms of $D_+$, equation ( 14 ).

$$D_- = -D_+ \frac{1 + \lambda_+ R_A C_R}{1 + \lambda_- R_L C_M} \qquad (\,14\,)$$

Equation ( 14 ) is then substituted into equation ( 12 ) to obtain an expression for $D_+$.

$$D_+ = \frac{V_0}{1 - \frac{1 + \lambda_+ R_A C_R}{1 + \lambda_- R_L C_M}} \qquad (\,15\,)$$

Finally, equation ( 15 ) is substituted into equation ( 14 ) to obtain and expression for $D_-$.

$$D_- = -D_+ \frac{1 + \lambda_+ R_A C_R}{1 + \lambda_- R_L C_M} \qquad (\,16\,)$$

Having determined the coefficients of our expression for $v_R(t)$, we can then obtain the final expression for our output voltage, $v_M(t)$, by substituting equation ( 11 ) into equation ( 7 ).

$$v_M(t) = (1 + R_A C_R \lambda_+)D_+ exp^{\lambda_+ t} + (1 + R_L C_M \lambda_-)D_- exp^{\lambda_- t} \qquad (\,17\,)$$

Note, a key limitation of this analysis is that in practice, charging the reservoir capacitor will also lead to some charging of membrane capacitor affecting the initial conditions we assumed previously. This will become worse as the voltage pulse used to charge the capacitor becomes wider. Hence, the analysis is best suited for simulating the response to short voltage pulses at the input which negligibly charge the membrane capacitor.

## S2. Sound Localisation

The component values of each of the three variants are detailed in the table below.

Table 1

| Sub-circuit | Component | Component Values of Each Circuit Variant | | |
| --- | --- | --- | --- | --- |
| | | A | B | C |
| N1 | $C_M$ | 1 μF | 1 μF | 1 μF |
| | $C_R$ | 1 μF | 1 μF | 1 μF |
| | $R_A$ | 1.42 $k\Omega$ | 2.18 $k\Omega$ | 2.7 $k\Omega$ |
| | $R_L$ | 1.94 $k\Omega$ | 5.34 $k\Omega$ | 8.89 $k\Omega$ |
| N2 | $C_M$ | 1 μF | 1 μF | 1 μF |
| | $C_R$ | 1 μF | 1 μF | 1 μF |
| | $R_A$ | 205 $\Omega$ | 196 $\Omega$ | 197 $\Omega$ |
| | $R_L$ | 140 $\Omega$ | 140 $\Omega$ | 132 $\Omega$ |
| P1 | $C_M$ | 1 μF | 1 μF | 1 μF |
| | $C_R$ | 1 μF | 1 μF | 1 μF |
| | $R_A$ | 13 $\Omega$ | 13 $\Omega$ | 13 $\Omega$ |
| | $R_L$ | 1 $k\Omega$ | 1 $k\Omega$ | 1 $k\Omega$ |

## S3. Neuron Imitation

The bursting neuron from figure 6 of the article had the structure as shown below. The component values for each of the sub-circuits are detailed in the Table 2.

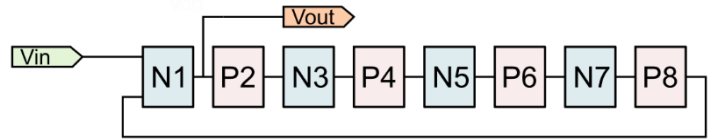

**Fig. 1 | Block diagram of a dendritic bursting neuron.** A bursting neuron is imitated by a ring of dendrite segments. The circuit for each block is documented in the main article. The component values for each sub-circuit

Table 2

| Sub-circuit | $R_A (\Omega)$ | $R_L (\Omega)$ | $C_R (nF)$ | $C_M (nF)$ |
| --- | --- | --- | --- | --- |
| | | Component Values | | |
| N1 | 220 | 1000 | 3.3 | 3.3 |
| P2 | 220 | 377 | 22 | 22 |
| N3 | 220 | 220 | 22 | 22 |
| P4 | 220 | 438 | 22 | 22 |
| N5 | 220 | 187 | 22 | 22 |
| P6 | 220 | 390 | 22 | 22 |
| N7 | 220 | 220 | 22 | 22 |
| P8 | 220 | 127 (1 spike) 231 (2 spikes) 251 (3 spikes) | 22 | 22 |